\documentclass[lettersize,journal]{IEEEtran}
\usepackage{amsmath,amsfonts}
\usepackage{algorithmic}
\usepackage{algorithm}
\usepackage{array}
\usepackage[caption=false,font=normalsize,labelfont=sf,textfont=sf]{subfig}
\usepackage{textcomp}
\usepackage{stfloats}
\usepackage{url}
\usepackage{verbatim}
\usepackage{graphicx}
\usepackage{cite}
\usepackage{color}
\usepackage{amsthm,amsmath,amssymb}
\usepackage{bbm}
\usepackage{dsfont}
\usepackage{booktabs}
\begin{document}

\title{PolyBuilding: Polygon Transformer for End-to-End Building Extraction}

\author{Yuan Hu, Zhibin Wang, Zhou Huang, Yu Liu
\thanks{Yuan Hu and Zhibin Wang are with the DAMO Academy, Alibaba Group, Beijing 100102, China (e-mail: lavender.hy@alibaba-inc.com; zhibin.waz@alibaba-inc.com).}
\thanks{Zhou Huang and Yu Liu are with the Institute of Remote Sensing and Geographic Information Systems, School of Earth and Space Sciences, Peking University, Beijing 100871, China. (e-mail: huangzhou@pku.edu.cn; yuliugis@pku.edu.cn).}
}



\maketitle

\begin{abstract}
We present PolyBuilding, a fully end-to-end polygon Transformer for building extraction.
PolyBuilding direct predicts vector representation of buildings from remote sensing images.
It builds upon an encoder-decoder transformer architecture and simultaneously outputs building bounding boxes and polygons.
Given a set of polygon queries, the model learns the relations among them and encodes context information from the image to predict the final set of building polygons with fixed vertex numbers.
Corner classification is performed to distinguish the building corners from the sampled points, which can be used to remove redundant vertices along the building walls during inference.
A 1-d non-maximum suppression (NMS) is further applied to reduce vertex redundancy near the building corners.
With the refinement operations, polygons with regular shapes and low complexity can be effectively obtained.
Comprehensive experiments are conducted on the CrowdAI dataset.
Quantitative and qualitative results show that our approach outperforms prior polygonal building extraction methods by a large margin. It also achieves a new state-of-the-art in terms of pixel-level coverage, instance-level precision and recall, and geometry-level properties (including contour regularity and polygon complexity).
\end{abstract}


\section{Introduction}
\IEEEPARstart{P}{olygonal}
building extraction from aerial and satellite imagery is of great significance in practical applications, such as cartography \cite{turker2015building}, 3D-city modeling \cite{lafarge2008automatic,sportouche2009building}, cadastral and topographic mapping \cite{sun2021building}, and so on.
Compared to the raster data format of building segmentations, the vector data structure of building polygons is more compact, scalable, and geolocation accurate. Moreover, vector data encode desired topological information that can support many topology operations in geographic information systems, such as proximity operations, network analysis, etc.

Building extraction has been a research hotspot for decades as the continuous increase of very high-resolution remote sensing imagery \cite{simonetto2005rectangular,ji2018fully,jung2021boundary,buildingvit}.
We consider that desired building predictions need to meet the following requirements. First, pixel-level coverage: the predictions should approximate the building footprints well at the pixel level. Second, instance-level detection: the model should detect all building instances accurately without missed and false detections. Third, geometry-level properties: the predicted buildings should have regular shapes, such as sharp corners and straight edges.
In addition, polygon complexity should be taken into account, i.e., low vertex redundancy is necessary for the generated building polygons.
As for the building extraction models, many prior methods consist of complex processing procedures for generating polygonal outputs. Some of the procedures can not be processed in parallel on GPU.
Thus, we consider that a desired polygonal building extraction model should have the following characteristics. First, efficient and parallelizable: parallelizable on the GPU is extremely important to guarantee efficient model training under the background of the ever-increasing amount of remote sensing imagery. Second, concise and effective: end-to-end models are preferred since they can reduce a lot of manual involvement.

Most previous attempts adopt segmentation-based methods for polygonal building extraction, followed by a series of post-processing procedures \cite{ffl,icpr}. The segmentation results often have great pixel-level coverage of the building footprints and achieve decent Intersection-over-Union (IoU) scores. However, they tend to be blob-like with rounded corners and irregular contours. These imperfections make the subsequent vectorization very difficult. In addition, they often need to train several separate models for segmentation, regularization, and vectorization \cite{icpr}, which are complex and inefficient. Errors are also accumulated throughout the whole pipeline.

\IEEEpubidadjcol

Another category of methods learns polygonal representations for building instances directly \cite{polyworld,aaai,polymapper, polymapper-follower,curvegcn-based}.
The generated buildings usually have more regular shapes than segmentation-based methods.
Different ideas are proposed for extracting polygonal buildings from the image.
For example, \cite{polymapper,polymapper-follower} first detect bounding boxes for the buildings using an object detection model (such as Faster R-CNN \cite{faster-rcnn}, FPN \cite{fpn}), and then within the boxes, building corners are predicted sequentially to form a polygon using a recurrent neural network (RNN, such as ConvLSTM \cite{convlstm}, ConvGRU \cite{convgru}). However, the CNN-RNN paradigm is not concise and difficult to train.
Some other methods \cite{curvegcn-based} replace the RNN part with a graph convolutional network (GCN) to predict the polygon vertices at one time instead of one at a time. Although the CNN-GCN paradigm is more efficient than the CNN-RNN one, they predict a fixed number of vertices for all building instances, which causes heavy vertex redundancy.
\cite{polyworld} tackles the task from another perspective. It directly detects all the building corners in the input images using a convolutional neural network (CNN) and then connects them in order according to a learned permutation matrix indicating the connections between vertices, which is more concise than previous methods. However, it often suffers from missed detections of building corners, even the whole building instances, since the vertices are extracted through vertex segmentation, which is sensitive to the occlusions of vegetation or shadows.

In this paper, we tackle the polygonal building extraction task leveraging a totally novel and fully end-to-end method --- PolyBuilding.
PolyBuilding is a transformer-based model, which predicts building bounding boxes and polygons simultaneously, which is different from previous methods \cite{polymapper,polymapper-follower,curvegcn-based} that predict bounding boxes first and then detect polygon vertices within the previous predicted boxes.
Transformers \cite{transformer} are initially developed for natural language processing tasks \cite{bert} and subsequently achieve notable success in computer vision \cite{vit}. DEtection TRansformer (DETR) \cite{detr} builds the first fully end-to-end object detector by leveraging the encoder-decoder transformer architecture. It directly predicts a set of object bounding boxes, which totally eliminates hand-crafted components, such as anchor generation, needed in previous detectors \cite{faster-rcnn, fpn, yolo, ssd}. 
Deformable DETR \cite{deformable-detr} further improves the convergence speed and detection performance of the original DETR using the proposed multi-scale deformable attention module.
Polygonal building extraction can be viewed as an extension of object detection by further regressing the coordinates of polygon vertices.
Inspired by the similarity of the two tasks, we propose PolyBuilding by extending Deformable DETR \cite{deformable-detr} to further predict building polygons besides the bounding boxes.
This is achieved by adding a polygon regression head and a corner classification head.
The polygon regression head regresses the coordinates of polygon vertices for a set of building instances.
All the building instances are transformed to have a fixed number of vertices using the proposed uniform sampling encoding scheme.
Thus, the corner classification head is designed to classify the vertices as building corners or sampled points for each building polygon, which is used to filter out redundant vertices along the building walls during inference.
In practice, a simple 1-d non-maximum suppression (NMS) is further applied to the predicted corner score sequence to remove redundant vertex near the building corners.
With the refinement operations, polygons with regular shapes and low complexity can be obtained.
The proposed PolyBuilding model inherits the advantages of Deformable DETR, which is fully end-to-end, more concise, and easy to train than previous CNN-RNN or CNN-GCN paradigm based methods \cite{polymapper,polymapper-follower,curvegcn}.

Extensive experiments are conducted on the CrowdAI dataset to evaluate the proposed PolyBuilding model.
Experimental results show that PolyBuilding can achieve remarkable performance in pixel-level coverage (measured by Intersection over Union (IoU)), instance-level detection (measured by Average Precision (AP) and Average Recall (AR)), and geometry-level properties (measured by Max Tangent Angle Error (MTA), N ratio, and complexity aware IoU (C-IoU)).

The contributions of this paper are three-fold, summarized as follows.
\begin{itemize}
    \item To the best of our knowledge, PolyBuilding is the first transformer-based polygonal building extraction model, which is concise and fully end-to-end trainable.
    \item Building polygons generated by PolyBuilding satisfy the requirements of great building coverage, high detection precision and recall, and great geometric properties (regular contours and low polygon complexity).
    \item PolyBuilding achieves a new state-of-the-art on CrowdAI dataset.
\end{itemize}

\section{Related work}
Polygonal building extraction methods can be mainly classified into two categories, namely segmentaion-based multi-stage methods and end-to-end direct prediction methods.

\subsection{Segmentation-based Polygonal Building Extraction}
Segmentation-based methods \cite{ffl,icpr,maskrcnn_regularization,asip,chen2021quantization} first predict the building probability maps using a semantic or instance segmentation network \cite{buildingvit,asfnet,bomsc,bfgcnet,multiscale,jung2021boundary,tian2021multiscale,liu2020multiscale,zhu2020map,wei2019toward} and then performs post-processing operations on the segmented building outlines to obtain the final vector format outputs.
For example, \cite{maskrcnn_regularization} applied Mask R-CNN \cite{maskrcnn} to produce building instance masks, and then employed a series of operations to regularize and vectorize the boundaries, including simplifying the initial polygons using the Ramer-Douglas-Peucker algorithm \cite{douglas1973algorithms}, and then refining the outputs using the Minimum Descriptor Length optimization \cite{mdl}. In \cite{icpr}, three models were sequentially adopted for building segmentation, boundary regularization, and polygonization. Specifically, a R2U-Net \cite{r2unet} was used to segment building footprints, and then a generative adversarial network (GAN) was utilized to regularize the boundaries; finally a convolutional neural network (CNN) was adapted to predicts building corners. The polygonization was accomplished by ordering the corners along the regularized contours. FFL (Frame Field Learning) \cite{ffl} achieves the state-of-the-art in this category of methods. It trained a deep segmentation model with an additional frame field supervision. The frame field objective not only improves the segmentation quality but also facilitates later polygonization.
In summary, these methods are usually composed of multiple models or processing steps, which are complex and involve too many hand-designed rules and hyperparameters. In addition, the quality of prior segmentation results heavily restricts the performance that the subsequent vectorization can achieve.

\subsection{End-to-end Polygonal Building Extraction}
\label{sec-relate-work-e2e}
End-to-end methods predict building polygons directly. Polygon-RNN \cite{polygonrnn} and Polygon-RNN++ \cite{polygonrnn++} are semi-automatic object annotation methods, which generate a polygon for each object instance using a Recurrent Neural Network (RNN) given the ground-truth bounding box.
PolyMapper \cite{polymapper} extended this architecture by adding a prepositive object detection module. It predicted building bounding boxes using the Feature Pyramid Network (FPN) architecture \cite{fpn} and then sequentially predicted the building corners using the ConvLSTM model \cite{convlstm} inside the extracted bounding boxes. \cite{polymapper-follower} further improved PolyMapper by upgrading the detection module as well as the RNN part. The CNN-RNN paradigm is inefficient to train and inference due to its complex architecture and the beam search procedure, respectively.
Curve-GCN \cite{curvegcn} is more efficient than the Polygon-RNN families \cite{polygonrnn,polygonrnn++} by predicting all polygon vertices simultaneously using a Graph Convolutional Network (GCN).
\cite{curvegcn-based} followed the Curve-GCN method \cite{curvegcn} to predict the building corners at one time and also added an object detection module before it. It is more efficient, but the limitation of the CNN-GCN paradigm is the vertex redundancy caused by the preset fixed vertex number.
Some other methods \cite{polyworld, aaai} predicted building polygons by performing corner detection as a segmentation task and then applied a GCN module to refine the coordinates of the formerly extracted vertices.
PolyWorld \cite{polyworld} adopted this method and further predicted a permutation matrix to encode the connection information between extracted vertices. Vertices were then connected according to the permutation matrix to form the final polygons. Although achieving great improvements on CrowdAI dataset \cite{crowdai}, PolyWorld suffers from miss detection of building corners caused by vegetation or shadow occlusion during the vertex segmentation procedure.
PolyBuilding proposed in this paper falls into this category of methods. It novelly leverages a transformer-based architecture to tackle the polygonal building extraction task.


\begin{figure*}[!t]
\centering
\includegraphics[width=17cm]{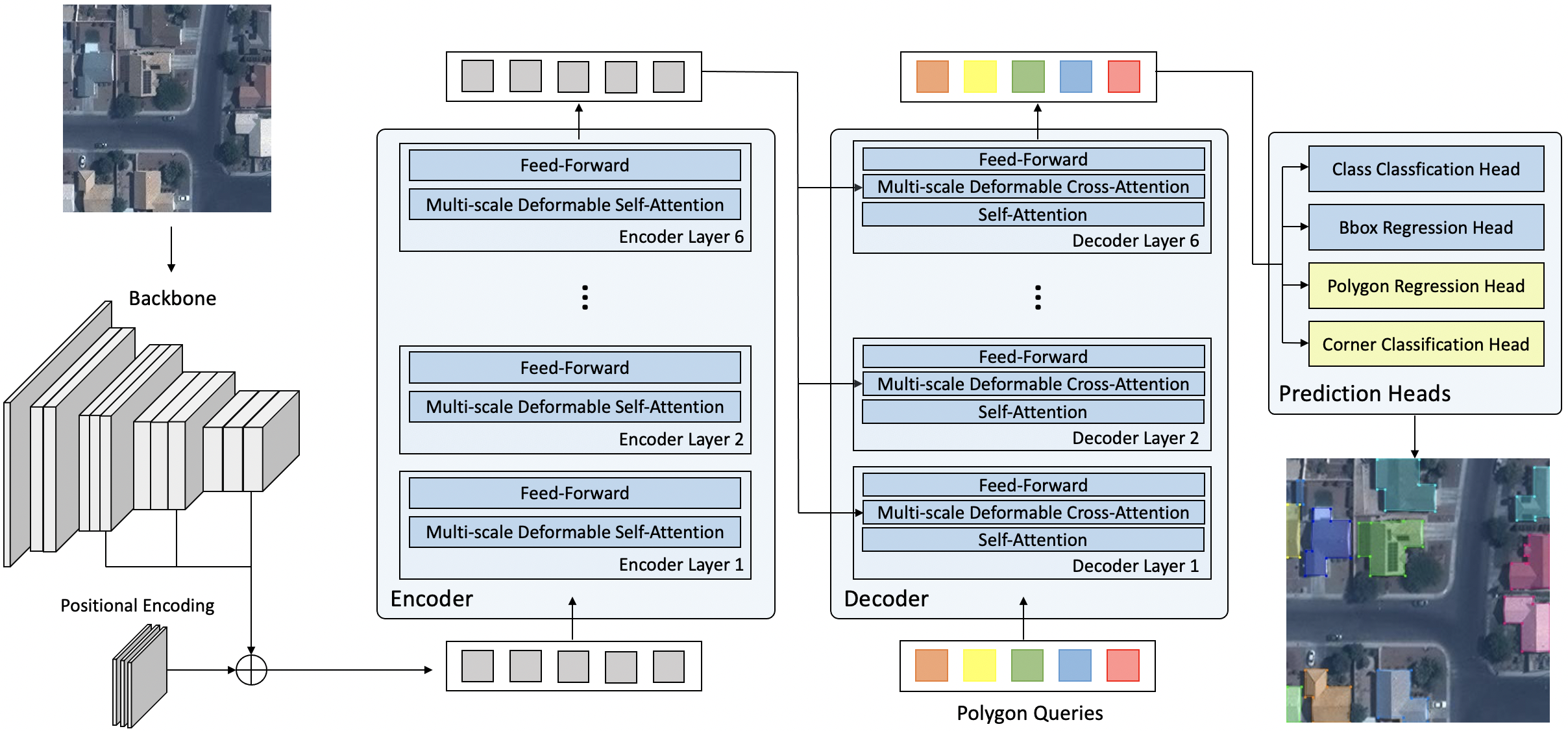}
\caption{Overall architecture of PolyBuilding. The input image is first fed into a backbone network to extract multi-scale features with lower spatial resolutions, which are then added with the positional embedding and flattened to be ready for entering the transformer. The transformer encoder takes as input the multi-scale features and performs multi-scale deformable self-attention to output enhanced features with long-range relation encoded. Randomly initialized polygon queries are then fed into the transformer decoder, in which self-attention is applied for the information interaction among the polygon queries and multi-scale deformable cross-attention is further employed to encode context information into the queries from the multi-scale features of the encoder outputs. Finally, prediction heads are utilized for predicting the class classification scores, bounding boxes, polygon coordinates, and corner classification scores for building instances, which are combined to generate the final building polygons.}
\label{fig-architecture}
\end{figure*}

\section{Methodology}
Following Deformable DETR \cite{deformable-detr}, PolyBuilding consists of similar components. In this section, we first introduce the overall architecture of PolyBuilding, and then elaborate on the building polygon encoding scheme for the generation of the polygon ground truth. Finally, we present the training strategies and loss functions.

\subsection{Overall Architecture}
The overall architecture of PolyBuilding is illustrated in Figure \ref{fig-architecture}. It contains three main components: a backbone, an encoder-decoder transformer, and task-specific prediction heads. In general, given the input remote sensing images, the backbone is first used to extract multi-scale features. Then a standard transformer architecture is exploited to further transform the feature maps to a set of polygon embedding features. Finally, prediction heads are adopted to generate building classification scores, bounding boxes, polygon coordinates, and corner classification scores, which are leveraged to generate final building polygons.

\subsubsection{Backbone}
Since self-attention in the transformer encoder is of high computational and memory complexities in processing imagery data, the self-attention computation is the quadratic complexity of the number of image pixels. Thus, it is conventional to first apply a backbone network to extract lower-resolution feature maps. The backbone network can be a Convolutional Neural Network (CNN) \cite{resnet} or a vision transformer\cite{vit,swin}. In this paper, we adopt ResNet-50 \cite{resnet} as the backbone.
We extract multi-scale features (i.e., C3, C4, and C5) from the last three stages of the ResNet, of which the resolutions are 1/8, 1/16, and 1/32 of the input image, respectively. We further extract a 1/64 scale feature by downsampling the 1/32 one with a $3\times3$ stride 2 convolution. The four-scale features are then transformed by a $1\times1$ convolution to output new ones with 256 channels, which are added with the positional embedding, and then flattened and concatenated to be ready for entering the transformer.

\subsubsection{Transformer}
The transformer consists of an encoder and a decoder, of which the main component is the multi-scale deformable attention module. Deformable attention module \cite{deformable-detr} can greatly improve the convergence speed and decrease memory cost by only attending to a small set of key sampling locations. It can also be easily extended to the multi-scale version to leverage multi-scale features.

Given a set of multi-scale feature maps $\{\boldsymbol{x}^l\}_{l=1}^L$ and the normalized coordinates of the reference point $\boldsymbol{p}_q$ for each query $q$, the multi-scale deformable attention module can be written as
\begin{multline}
    \text{MSDeformAttn}(q, \boldsymbol{p}_q, \{\boldsymbol{x}^l\}_{l=1}^L) = \\
    \sum_{m=1}^M \boldsymbol{W}_m [\sum_{l=1}^L \sum_{k=1}^K A_{mlqk} \cdot \boldsymbol{W}_m^{'} \boldsymbol{x}^l (\phi_l(\boldsymbol{p}_q) + \Delta \boldsymbol{p}_{mlqk})]
\end{multline}
where $m$, $l$, and $k$ index the attention head, feature level, and sampling points, respectively. $\boldsymbol{W}_m$ and $\boldsymbol{W}_m^{'}$ are learnable weight matrices. $A_{mqk}$ denotes the attention weight for query $q$. $\phi_l(\boldsymbol{p}_q)$ re-scales the normalized coordinates $\boldsymbol{p}_q$ to the $l$-th level feature map. $\Delta \boldsymbol{p}_{mlqk}$ denotes the learned sampling offset. $\phi_l(\boldsymbol{p}_q) + \Delta \boldsymbol{p}_{mlqk}$ achieves the sampling location of query $q$ on the feature map $\boldsymbol{x}^l$.

The transformer encoder contains 6 identical encoder layers. Each is composed of a multi-scale deformable self-attention layer and a feed-forward network. 
The query elements in the multi-scale deformable self-attention layer are pixels of the multi-scale features, and the key elements are $LK$ points sampled from the multi-scale features instead of all spatial locations, thus it can reduce computational complexity and accelerate convergence speed.
In general, the transformer encoder takes as input the multi-scale ResNet feature maps and output enhanced ones with long-range relation encoded.

The transformer decoder is also composed of 6 identical decoder layers. Each has a self-attention layer, a multi-scale deformable cross-attention layer, and a feed-forward network. We initialize $N$ polygon queries randomly as the input of the decoder, which means that we predict $N$ building polygons for each image (some of them may be no objects). In the self-attention modules, both the query and key elements are polygon queries, and they interact with each other to capture their relations. In the cross-attention module, the query elements are kept the same, but the key elements are feature maps from the encoder output, thus the polygon queries can encode context information from the feature maps through cross attention. Similarly, the queries only attend to $LK$ points in the multi-scale features. Finally, the decoder outputs $N$ feature embeddings corresponding to $N$ building polygons.

\subsubsection{Prediction heads}
In order to obtain building polygon predictions, we add two additional prediction heads (i.e., the polygon regression head and the corner classification head) besides the original class classification head and bounding box regression head in Deformable DETR \cite{deformable-detr}.
The class classification head predicts building classification confidence scores for all the $N$ polygons, specifying the buildings and no objects. The bounding box regression head regresses the normalized center coordinates as well as the height and width of the bounding boxes for all the $N$ objects. The polygon regression head is designed to further regress the coordinates of $M$ vertices for each polygon.
Since different building instances have various shapes and a different number of corners or annotated points, but the model inputs or outputs must be of the same shape.
To tackle this issue, we first unify the vertex number for all the annotated building polygons using a building polygon encoding scheme, which will be introduced in the following sub-section. However, keeping all the buildings with the same number of vertices would cause vertex redundancy, especially for simple buildings.
In order to control the polygon complexity and reduce vertex redundancy, we further develop the corner classification head to distinguish the corners among the $M$ vertices for each building instance. During inference, clean building polygons with low complexity can be obtained by combining the outputs of the polygon regression head and the corner classification head. 
Specifically, most vertices located in the building walls can be filtered according to the corner classification confidence scores.
However, in practice, vertices near the corners may be reserved due to relatively high scores.
Thus, a simple 1-d non-maximum suppression (NMS) along the corner score sequence is adopted to further reduce the vertex redundancy near the corners.
With the polygon refinement operations, clean polygons with low complexity are obtained.
Regarding the architectures of the prediction heads, the class classification head is a simple linear projection layer. The other three heads have identical architectures, i.e., 3-layer feed-forward networks with ReLU activation functions.

\begin{figure}[!t]
\centering
\includegraphics[width=8.5cm]{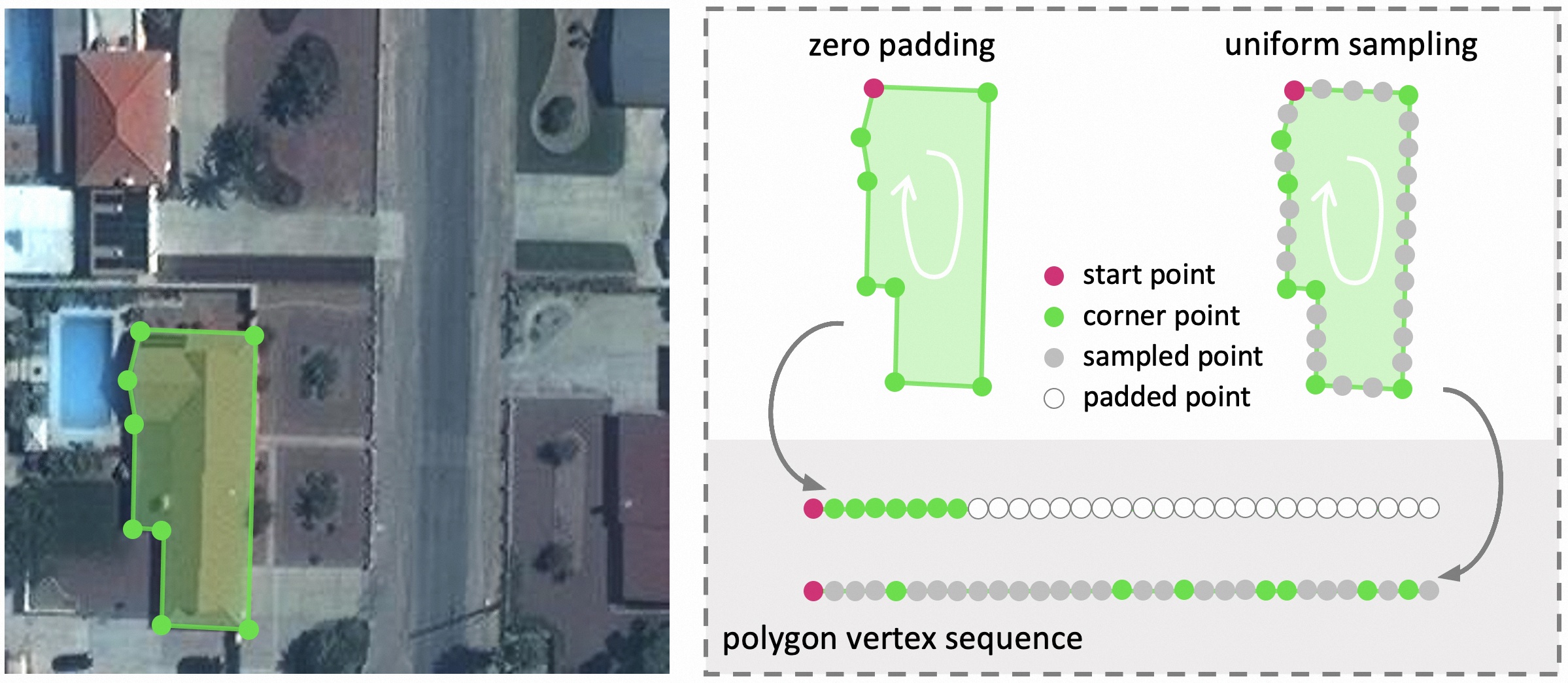}
\caption{Illustration of building polygon encoding schemes. The ground truth annotated points of the building in the lower left part of the image are labeled using green points. To generate a vertex sequence with a fixed number ($M$) of points, the zero padding scheme simply pads zeros at the end of the sequence, and uniform sampling scheme uniformly samples $M$ points along the building boundaries. Both of them rearrange the vertex sequence in clockwise order and start the sequence from the top extreme point. The start points, corner points, sampled points, and padded points are marked in red, green, gray, and white colors, respectively.}
\label{fig-polygon-encoding}
\end{figure}

\subsection{Building Polygon Encoding}
\label{sec-polygon-encoding-scheme}
Since the input polygon queries and output polygon coordinates for different buildings must be the same shape in the architecture of the transformer, we adopt building polygon encoding to unify the vertex number for all the annotated buildings. There are two feasible encoding schemes, namely uniform sampling and zero padding, as illustrated in Figure \ref{fig-polygon-encoding}.

\subsubsection{Uniform sampling}
We uniformly sample $M$ vertices along the ground truth building boundaries in the clockwise order starting from the top extreme point. Note that uniform sampling may miss the corner points and cause round contours near the corners. We tackle this issue by solving a linear sum assignment problem.
Specifically, we create the cost matrix according to the Euclidean distances between the annotated points and the uniform sampled points and then replace the sampled points with the annotated points with minimal cost according to the cost matrix.
In this way, we reserve all the annotated points and guarantee the corners are orthogonal.
In addition, we generate a score sequence corresponding to the vertex sequence for each building instance. The score sequence also has a length of $M$. We mark the corners (annotated points) as $1$ and the remaining sampled points as $0$, which can be used to filter out redundant vertices during inference. The generated ground truth vertex coordinate sequence and score sequence are used to supervise the polygon regression head and corner classification head, respectively.

\subsubsection{Zero padding}
It is much easier to generate the zero-padded ground truth. We first re-adjust the order of the polygon vertices in clockwise order and start the sequence from the top extreme point. After that, we add zeros at the end of the vertex sequence with vertices less than $M$.
If an annotated polygon has vertices greater than $M$, we will simplify the polygon by deleting the top $k$ shortest edges to ensure the vertex number exactly equals $M$.
The simplification would harm the polygon topology, thus we set $M=96$ to cover most building polygons.
For the score sequence, we mark the original annotated points as $1$ and the remaining padded elements as $0$ to indicate the padded parts. We compare the performance of training with the uniform sampled and zero-padded ground truth in Section \ref{sec-ablation-study}.

\subsection{Training}
\subsubsection{Training strategy}
The whole model can be end-to-end trained with both transformed ground truth. However, for the uniform sampling scheme, we design a two-phase training strategy.
In the first phase, we use all-ones ground truth to supervise the corner classification head, i.e., we do not distinguish the corners and sampled points for the building instances. In this way, the corner classification for the polygons is consistent with the class classification for the bounding boxes, thus both of them are trained for classifying the objects as buildings or no objects.
This simplifies the model objectives and facilitates the optimization for the location regression of the bounding boxes and polygons. In the second phase, we finetune the model using the score sequence ground truth that marks the corners with ones and sampled points with zeros for the corner classification head, which makes the optimization in this training stage focus on distinguishing corners out of the $M$ vertices for each polygon. The two-phase training strategy can achieve better convergency by decomposing the objectives of coordinate learning and corner classification. We compare the performance of end-to-end training and the proposed two-phase training in Section \ref{sec-ablation-study}.

\subsubsection{Training losses}
PolyBuilding outputs a fixed number of $N$ predictions for all images. Therefore, we need to find an optimal bipartite matching between the predicted and the ground truth building instances before calculating the losses. The matching cost $\mathcal{C}_{match}$ is composed of the class classification part and the bounding box prediction part following \cite{deformable-detr}:
\begin{multline}
\label{eq-matching-cost}
    \mathcal{C}_{match}(y_i, \hat{y}_{\sigma(i)})) =  \lambda_{cls} \mathcal{L}_{FL}(\hat{p}_{\sigma(i)}) \\
    + \lambda_{iou} \mathcal{L}_{iou}(b_i, \hat{b}_{\sigma(i)}) + \lambda_{L1} \| b_i-\hat{b}_{\sigma(i)})\|_1
\end{multline}
where $\mathcal{C}_{match}(y_i, \hat{y}_{\sigma(i)}))$ denotes the matching cost between the ground truth $y_i$ and the prediction $\hat{y}_{\sigma(i)}$ with index $\sigma(i)$. $\hat{p}_{\sigma(i)}$ denotes the probability of $\hat{y}_{\sigma(i)}$ being a building. Focal loss \cite{focalloss} is adopted to calculate the classification loss. $b_i$ and $\hat{b}_{\sigma(i)}$ denote the center coordinates and height and width of the ground truth and predicted boxes, respectively. The second and third term in equation (\ref{eq-matching-cost}) denotes the generalized IoU loss \cite{giou} and L1 loss between the ground truth and predicted bounding boxes, respectively. $\lambda_{cls}$, $\lambda_{iou}$, and $\lambda_{L1}$ are balancing hyperparameters.  An optimal matching is assigned by minimizing the matching cost.

After that, losses can be calculated according to the matched ground truth and prediction pairs. The overall loss function of our PolyBuilding model can be written as:
\begin{equation}
\label{eq-overall-loss}
    \mathcal{L} = \lambda_{cls} \mathcal{L}_{cls} + \mathcal{L}_{bbox} + \lambda_{poly} \mathcal{L}_{poly} + \lambda_{cnr} \mathcal{L}_{cnr}
\end{equation}
where $\mathcal{L}_{cls}$, $\mathcal{L}_{bbox}$, $\mathcal{L}_{poly}$, and $\mathcal{L}_{cnr}$ are losses corresponding to the four prediction heads. $\lambda_{cls}$, $\lambda_{poly}$, and $\lambda_{cnr}$ are balancing hyperparameters. $\mathcal{L}_{bbox}$ is composed of two parts, thus the balancing hyperparameters for $\mathcal{L}_{bbox}$ are displayed in Equation (\ref{eq-bbox-loss}).

We adopt focal loss as the class classification loss $\mathcal{L}_{cls}$ for building instances:
\begin{multline}
    \mathcal{L}_{cls} = \sum_{i=1}^N( -\mathds{1}_{\{y_i\neq\varnothing\}} \alpha(1-\hat{p}_{\hat{\sigma}(i)})^\gamma \log(\hat{p}_{\hat{\sigma}(i)}) \\
    -\mathds{1}_{\{y_i=\varnothing\}} (1-\alpha)(\hat{p}_{\hat{\sigma}(i)})^\gamma \log(1-\hat{p}_{\hat{\sigma}(i)}))
\end{multline}
where $\varnothing$ denotes no object, $\hat{\sigma}(i)$ denotes the index of optimal bipartite matching. $\alpha$ and $\gamma$ are hyperparameters of focal loss.

Bounding box loss $\mathcal{L}_{bbox}$ is a linear combination of the generalized IoU loss \cite{giou} and the L1 loss:
\begin{equation}
\label{eq-bbox-loss}
    \mathcal{L}_{bbox} = \sum_{i=1}^N(\lambda_{iou} \mathcal{L}_{iou}(b_i, \hat{b}_{\hat{\sigma}(i)}) + \lambda_{L1} \| b_i-\hat{b}_{\hat{\sigma}(i)})\|_1)
\end{equation}

For building polygon coordinate regression, we also adopt L1 loss:
\begin{equation}
    \mathcal{L}_{poly} = \mathds{1}_{\{y_i\neq\varnothing\}} \sum_{i=1}^N(\| m_i-\hat{m}_{\hat{\sigma}(i)})\|_1)
\end{equation}
where $m_i, \hat{m}_{\hat{\sigma}(i)} \in \mathcal{R}^{M\times2}$ denotes the coordinate sequence of a ground truth building polygon and the corresponding matched prediction polygon, respectively.

Cross entropy loss is used as the corner classification loss:
\begin{equation}
    \mathcal{L}_{cnr} = \sum_{i=1}^N(-c_i\log(\hat{c}_{\hat{\sigma}(i)}) - (1-c_i)\log(1-\hat{c}_{\hat{\sigma}(i)}))
\end{equation}
where $c_i \in [0, 1]^M$ denotes the ground truth class label for the polygon vertices. $\hat{c}_{\hat{\sigma}(i)}$ denotes the predicted probabilities of the vertices being corners.

In addition to the model's final outputs, we add intermediate supervision for each decoder layer output. Thus, the final loss is the summation of the losses for all the $L$ decoder layers:
\begin{equation}
    \mathcal{L}_{final} = \sum_{i=1}^L \mathcal{L}_i,
\end{equation}
where each $\mathcal{L}_i$ has the same terms as in equation (\ref{eq-overall-loss}).

\section{Experimental settings and dataset}
\subsection{Dataset}
We evaluate our proposed method on the CrowdAI Mapping Challenge dataset \cite{crowdai}. CrowdAI dataset contains more than $340$k satellite images, in which $280,741$ tiles are used for training and $60,317$ tiles for testing. Each tile is a $300 \times 300$ pixel RBG image, and its corresponding annotation is in MS COCO format \cite{coco}. We upsample the images to $320 \times 320$ pixels for training and inference. Images and predicted polygon coordinates are rescaled back to the original image size for reporting the results.

\subsection{Evaluation Metrics}
We adopt three categories of metrics to evaluate the performance of PolyBuilding, i.e., pixel-level, instance-level, and geometry-level metrics.
For the pixel-level metric, we use Intersection over Union (IoU) to measure pixel-level coverage.
For instance-level metrics, Average Precision (AP) and Average Recall (AR) are adopted to assess the performance of building instance detection.
Based on different IoU thresholds, we calculate $AP$, $AP_{50}$, $AP_{75}$, which denotes average precision under the IoU thresholds of $0.5$ to $0.95$ (with a interval of 0.05), $0.5$, and $0.75$, respectively. $AR$, $AR_{50}$, $AR_{75}$ are computed in the similar manner.

In order to evaluate the geometric properties (such as contour regularity and polygon complexity) of the generated building polygons, we also compute geometry-level metrics, including Max Tangent Angle Error (MTA) \cite{ffl}, N ratio \cite{polyworld}, and complexity aware IoU (C-IoU) \cite{polyworld}. MTA compares the tangent angles between predicted and ground truth polygons so as to penalize irregular contours not aligned with the ground truth. It is the lower, the better. N ratio is the ratio between the vertex number of predicted and ground truth polygons, which is used to assess the polygon complexity. Predicted polygons with redundant vertices would have a ratio greater than $1$, and oversimplified polygons would get a ratio smaller than $1$, thus, the closer it is to $1$, the better.
In addition, we apply C-IoU \cite{polyworld} for evaluation considering both the pixel-level coverage and polygon complexity. C-IoU is defined as:
\begin{multline}
    \text{C-IoU}(A,\hat{A}) = (1-\text{RD}(N_A, N_{\hat{A}})) \cdot \text{IoU}(A, \hat{A}) \\
    \text{RD}(N_A, N_{\hat{A}}) = \frac{|N_A - N_{\hat{A}}|}{N_A + N_{\hat{A}}}
\end{multline}
where $A$ and $\hat{A}$ denote ground truth and predicted polygon mask, respectively. $N_A$ and $N_{\hat{A}}$ denote the vertex number of ground truth and predicted polygons, respectively. C-IoU is an IoU metric weighted by the vertex difference between ground truth and predicted polygons. Usually, predicted polygons with more vertices would have better coverage and higher IoU.
However, this metric favor a balance between coverage accuracy and polygon complexity. A prediction with both great coverage accuracy and proper polygon complexity is able to achieve a high C-IoU score.

\subsection{Experimental Details}
We set the uniform vertex number $M$ to $96$ to cover most annotated polygons. We set hyperparameters in equation (\ref{eq-overall-loss}) and (\ref{eq-bbox-loss}) as $\lambda_{cls}=2$, $\lambda_{poly}=5$, $\lambda_{cnr}=1$, $\lambda_{iou}=2$, $\lambda_{L1}=5$. Other hyper-parameter settings mainly follow Deformable DETR \cite{deformable-detr}. Unless otherwise specified, we train our model for $50$ epochs in ablation studies. The step learning rate policy is used with an initial learning rate $2\times10^{-4}$ and decayed by a factor of $0.1$ at the $40$-th epoch. When comparing with the state-of-the-arts, we train our model for $150$ epochs using the two-phase training strategy. In the first phase, PolyBuilding is trained for $100$ epochs with an initial learning rate $2\times10^{-4}$ that decays twice at the $40$-th and $80$-th epoch. In the second phase, the model is further finetuned for $50$ epochs with an initial learning rate $2\times10^{-4}$ that decays at $40$-th epoch. All experiments are optimized by AdamW using PyTorch and performed using 8 NVIDIA Tesla V100(32G) GPUs with batch size $48$.

\section{Experimental results}

\begin{table*}[!t]
\caption{Comparison Experiments Between Uniform Sampling and Zero Padding Building Polygon Encoding Schemes on CrowdAI Dataset. $\uparrow$ Denotes the Higher the Better, and $\downarrow$ Denotes the Lower the Better. N Ratio is the Closer to $1$, the Better. \label{tab-polygon-encoding}}
\centering
\begin{tabular}{c c c c c c c c c c c}
\toprule
Polygon encoding & AP $\uparrow$ & AP50 $\uparrow$ & AP75 $\uparrow$ & AR $\uparrow$ & AR50 $\uparrow$ & AR75 $\uparrow$ & IoU $\uparrow$ & MTA $\downarrow$ & N ratio & C-IoU $\uparrow$ \\
\midrule
zero padding & 59.6 & 92.3 & 68.9 & 68.6 & 94.7 & 79.4 & 83.7 & 36.0 & 0.93 & 78.9 \\
uniform sampling & \textbf{72.1} & \textbf{94.2} & \textbf{85.3} & \textbf{78.2} & \textbf{95.8} & \textbf{89.2} & \textbf{88.1} & \textbf{35.6} & \textbf{1.01} & \textbf{80.3} \\
\bottomrule
\end{tabular}
\end{table*}

\begin{table*}[!t]
\caption{Comparison Experiments Between End-to-end and Two-phase Training Strategies on CrowdAI Dataset. $\uparrow$ Denotes the Higher the Better, and $\downarrow$ Denotes the Lower the Better. N Ratio is the Closer to $1$, the Better. \label{tab-training-strategy}}
\centering
\begin{tabular}{c c c c c c c c c c c}
\toprule
Training strategy & AP $\uparrow$ & AP50 $\uparrow$ & AP75 $\uparrow$ & AR $\uparrow$ & AR50 $\uparrow$ & AR75 $\uparrow$ & IoU $\uparrow$ & MTA $\downarrow$ & N ratio & C-IoU $\uparrow$ \\
\midrule
end-to-end & 73.5 & 95.2 & 86.2 & 79.7 & 96.3 & 90.2 & 89.5 & 34.7 & 0.97 & 81.7 \\
two-phase & \textbf{78.7} & \textbf{96.3} & \textbf{89.2} & \textbf{84.2} & \textbf{97.3} & \textbf{92.9} & \textbf{94.0} & \textbf{32.4} & \textbf{0.99} & \textbf{88.6} \\
\bottomrule
\end{tabular}
\end{table*}

\subsection{Ablation Study} \label{sec-ablation-study}
In this subsection, we conduct ablation study to investigate the effectiveness of the components of the proposed PolyBuilding method, including the building polygon encoding schemes, the two-phase training strategy, and the polygon refinement operations.

We first evaluate the effect of the two building polygon encoding schemes, i.e., uniform sampling and zero padding.
Predictions trained with the zero-padded ground truth can be transferred to clean polygons by simply filtering out the points with low corner classification confidence scores in the tail of the vertex sequence. We empirically set the threshold to $0.1$. For uniform sampling, a NMS operation along the corner score sequence is further applied to remove redundant vertices near the building corners. Comparison results are shown in Table \ref{tab-polygon-encoding}. 
As can be seen, the uniform sampling scheme surpasses the zero padding scheme greatly in all evaluation metrics, especially the instance-level and pixel-level metrics. For example, uniform sampling is $12.5\%$, $9.6\%$, and $4.4\%$ higher than zero padding on AP, AR, and IoU, respectively. For geometry-level metrics, uniform sampling also achieves better performance ($35.6$ versus $36.0$ for MTA, $1.01$ versus $0.93$ for N ratio, and $80.3$ versus $78.9$ for C-IoU).
The reason may be that uniform sampled ground truth consists of dense vertices that provide rich location supervision information, making it easier to optimize the polygon coordinates. Nevertheless, zero-padded ground truth is too sparse to provide adequate supervision signals, thus harming the coordinates learning. 
The comparison results demonstrate the superiority of the uniform sampling encoding scheme.
Thus, we adopt it in the following experiments.

We then verify the effectiveness of the proposed two-phase training strategy for PolyBuilding. Table \ref{tab-training-strategy} shows the comparison results between the two-phase and end-to-end training strategies. Both the experiments are trained for $150$ epochs with batch size $48$. For end-to-end training, the supervision signal for the corner classification head is the ground truth with corners marked as $1$ and sampling points marked as $0$.
For the two-phase training, the ground truth for the first $100$ epochs and last $50$ epochs are all ones labels and labels that distinguish corners and sampling points, respectively.
As shown in Table \ref{tab-training-strategy}, the two-phase training strategy outperforms the end-to-end one in all evaluation metrics.
Specifically, regarding the instance-level and pixel-level metrics, the two-phase strategy achieves $3.7\%$, $3.8\%$, and $3.6\%$ gains in terms of AP, AR, and IoU, respectively. Furthermore, by observing AP50 and AP75 as well as AR50 and AR75, we can find that the main performance gains come from AP/AR  with higher IoU thresholds, demonstrating that the two-phase training strategy facilitates better optimization for vertex locations, thus producing more accurate polygon masks. Regarding geometric performance, the two-phase strategy also produces more regular polygon outputs with significantly lower MTA values ($32.4$ vs. $34.7$) and better polygon complexity similarity ($0.99$ vs. $0.97$). In addition, the two-phase strategy also boosts C-IoU value significantly from $81.7$ to $88.6$.
In summary, the results verify that the proposed two-phase training strategy can facilitate polygon learning with better pixel coverage, instance detection, and superior geometric performance.
We claim that this is achieved by decomposing the objectives of coordinate regression and corner classification and tackling them separately. In the first training phase, by supervising the corner classification head with all ones labels for $M$ polygon points, we unify the objectives of the corner classification head and the class classification head, i.e. both of them optimize for correctly classifying the true building instances out of the $N$ queries. This makes the optimization of the first training stage focus on the location regressions of the bounding boxes and polygons. On this basis, in the second training phase, the model can be easily fine-tuned for corner classification.

\begin{figure*}[!t]
\centering
\includegraphics[width=0.9\linewidth]{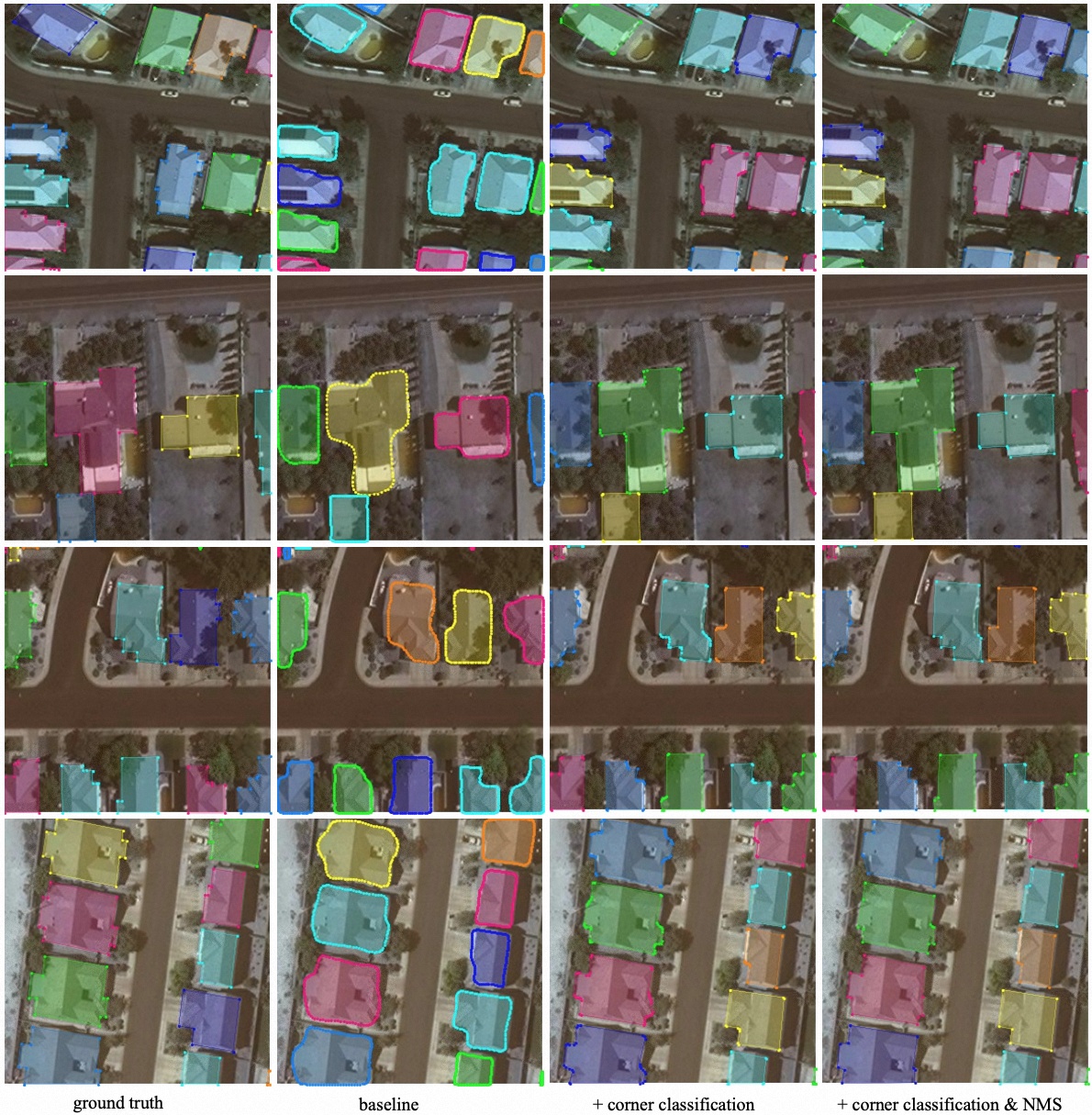}
\caption{Qualitative comparison of the experiments in Table \ref{tab-nms}. The first column shows the ground truth. The second to fourth columns show the predicted polygons with $M=96$ points, filtered by corner classification scores and 1-d NMS operation, respectively. The corner classification scores can effectively filter out redundant vertices along the building walls. NMS operation can further reduce redundancy near the corners.}
\label{fig-nms}
\end{figure*}

\begin{table}[!t]
\caption{The Influence of Corner Classification Filtering and NMS Filtering on the Geometric Performance of Generated Polygons. $\uparrow$ Denotes the Higher the Better, and $\downarrow$ Denotes the Lower the Better. N Ratio is the Closer to $1$, the Better. \label{tab-nms}}
\centering
\begin{tabular}{c c c c c c}
\toprule
Corner classification & NMS & MTA $\downarrow$ & N ratio & C-IoU $\uparrow$ \\
\midrule
& & 51.1 & 11.79 & 15.4 \\
\checkmark & & 43.4 & 2.10 & 61.3 \\
\checkmark & \checkmark & \textbf{32.4} & \textbf{0.97} & \textbf{85.6} \\
\bottomrule
\end{tabular}
\end{table}

Finally, we study the influence of polygon refinement operations on the geometric properties of the generated polygons of PolyBuilding. As mentioned before, predictions trained with uniform sampled ground truth need to be further processed to reduce vertex redundancy.
As shown in Table \ref{tab-nms} Row 1 and Row 2, adding corner classification improves polygon performance on all the geometry-related evaluation metrics. Specifically, it decreases MTA significantly from $51.1$ to $43.4$. This is because redundant vertices along the building walls would cause irregular non-linear contours and can be easily filtered out according to the corner classification scores. In this way, irregular contours composed of many points are simplified to straight lines, thus improving the boundary regularity and decreasing the MTA value. In addition, the N ratio value decreases from $11.79$ to $2.10$, which indicates that most redundant points have been removed. However, the value is still greater than $1$, meaning redundancy still exists. More notably, C-IoU is improved remarkably from $15.4$ to $61.3$, demonstrating the reduction of polygon complexity.
Row 2 and Row 3 in Table \ref{tab-nms} show the results of further applying NMS operation on the predicted polygons. All three metrics are further improved significantly. Specifically, MTA decreases from $43.4$ to $32.4$, indicating that the contour regularity of generated polygons is further improved. N ratio decreases from $2.10$ to $0.97$, indicating that the generated polygons transfer from having redundant points to being very similar to the ground truth. C-IoU also increases significantly from $61.3$ to $85.6$. 
We visualize the predictions with and without the refinement operations for intuitive comparison in Figure \ref{fig-nms}.
As shown in the second column of Figure \ref{fig-nms}, all the polygons are composed of $M=96$ points, resulting in non-linear boundaries for building walls and rounded curves for building corners. This explains the poor scores for MTA, N ratio, and C-IoU.
As shown in the third column of Figure \ref{fig-nms}, most redundant points along the building walls are filtered effectively by taking advantage of the corner classification scores. However, as can be seen, points near the building corner remain due to the relatively high confidence scores, consistent with the results in Table \ref{tab-nms}. The last column in Figure \ref{fig-nms} shows the final refined results. By adopting NMS along the score sequence, points with the highest confidence scores are reserved, and redundant points near the corners are removed. The results verify the effectiveness of our designed corner classification head and 1-d NMS operation on the generation of polygons with low complexity and high contour regularity.



\subsection{Comparison of State-of-the-Art Methods}

\begin{table*}[!t]
\caption{Comparison with State-of-the-arts on CrowdAI Dataset. $\uparrow$ Denotes the Higher the Better, and $\downarrow$ Denotes the Lower the Better. N Ratio is the Closer to $1$, the Better. \label{tab-sota}}
\centering
\begin{tabular}{c c c c c c c c c c c}
\toprule
Method & AP $\uparrow$ & AP50 $\uparrow$ & AP75 $\uparrow$ & AR $\uparrow$ & AR50 $\uparrow$ & AR75 $\uparrow$ & IoU $\uparrow$ & MTA $\downarrow$ & N ratio & C-IoU $\uparrow$ \\
\midrule
Mask R-CNN \cite{maskrcnn} & 41.9 & 67.5 & 48.8 & 47.6 & 70.8 & 55.5 & - & - & - & - \\
PANet \cite{panet} & 50.7 & 73.9 & 62.6 & 54.4 & 74.5 & 65.2 & - & - & - & - \\
\midrule
PolyMapper \cite{polymapper} & 55.7 & 86 & 65.1 & 62.1 & 88.6 & 71.4 & - & - & - & - \\
FFL \cite{ffl} & 60.9 & 87.4 & 70.4 & 64.5 & 89.2 & 73.4 & 84.4 & 33.5 & 1.13 & 74 \\
PolyWorld \cite{polyworld} & 63.3 & 88.6 & 70.5 & 75.4 & 93.5 & 83.1 & 91.3 & 32.9 & 0.93 & 88.2 \\
W. Li \textit{et al.} \cite{aaai} & 73.8 & 92 & 81.9 & 72.6 & 90.5 & 80.7 & - & - & - & - \\
\midrule
PolyBuilding (ours) & \textbf{78.7} & \textbf{96.3} & \textbf{89.2} & \textbf{84.2} & \textbf{97.3} & \textbf{92.9} & \textbf{94.0} & \textbf{32.4} & \textbf{0.99} & \textbf{88.6} \\
\bottomrule
\end{tabular}
\end{table*}

We compare the performance of the proposed PolyBuilding with state-of-the-art methods on the CrowdAI dataset in terms of pixel-level, instance-level, and geometry-level metrics. We compare with two building instance segmentation methods (Mask R-CNN \cite{maskrcnn} and PANet \cite{panet}) and four polygonal building extraction methods (PolyMapper \cite{polymapper}, FFL \cite{ffl}, PolyWorld \cite{polyworld}, and W. Li \textit{et al.} \cite{aaai}). Table \ref{tab-sota} shows the quantitative comparison results.

As can be seen in Table \ref{tab-sota},
PolyBuilding outperforms all the instance segmentation models and prior polygon-based models by a large margin in terms of instance-level metrics. Specifically, the AP and AR of the proposed PolyBuilding are $23.1\%$/$22.1\%$, $17.8\%$/$19.7\%$, $15.4\%$/$8.8\%$, and $4.9\%$/$11.6\%$ higher than PolyMapper, FFL, PolyWorld, and W. Li \textit{et al.}, respectively. Especially, the advantage of PolyBuilding is more significant as the IoU threshold increases. For example, PolyBuilding outperforms PolyWorld by $18.7\%$/$7.7\%$ for AP75/AP50 and $9.8\%$/$3.8\%$ for AR75/AR50.
The results demonstrate the capability of our model in generating building instances with high precision and recall, as well as accurate locations.
In addition, for pixel-level mask coverage, PolyBuilding also achieves the highest IoU value ($94.0\%$) among the compared methods, which is $9.6\%$ higher than FFL and $2.7\%$ higher than PolyWorld.
Regarding geometric performance, our model performs best on polygon contour regularity indicated by the MTA metric. Specifically, PolyBuilding achieves $32.4$ in terms of MTA, which is $3.4\%$ and $1.1\%$ higher than FFL and PolyWorld, respectively.
For polygon complexity, FFL generates $13\%$ more vertices than ground truth, and PolyWorld generates $7\%$ fewer vertices, measured by the N ratio metric. While PolyBuilding has the highest similarity to the ground truth with N ratio equals $0.99$.
For the C-IoU metric, PolyBuilding also outperforms FFL and PolyWorld, due to its superior pixel coverage and polygon complexity.
Overall, the comparison results verify the superiority and effectiveness of our proposed PolyBuilding model from the perspective of pixel-level coverage, instance-level precision and recall, and geometry-level properties.

\begin{figure*}[!t]
\centering
\includegraphics[width=18cm]{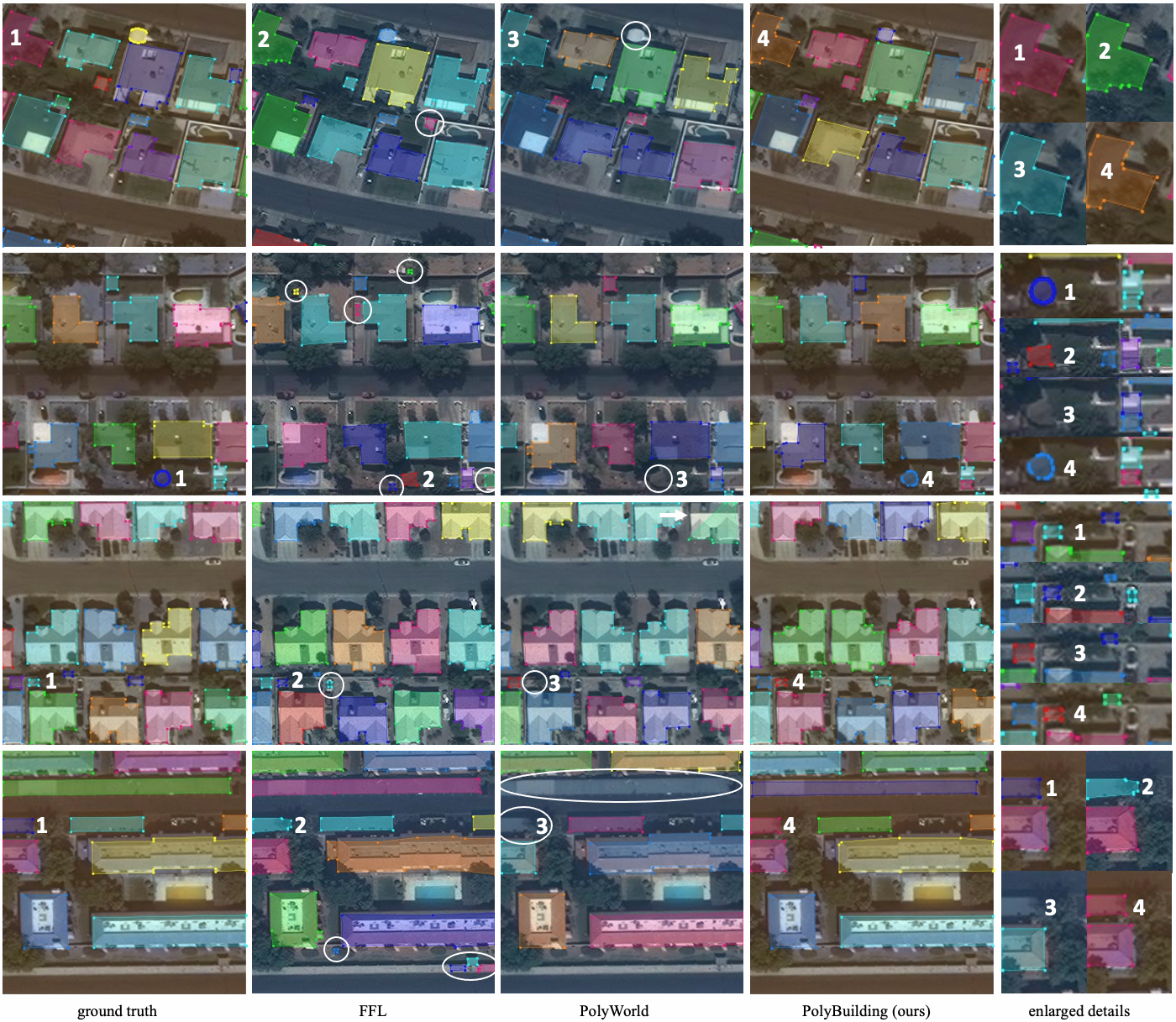}
\caption{Qualitative comparison on CrowdAI dataset among ground truth, FFL, PolyWorld, and PolyBuilding (ordering from the first column to fourth column in the figure). The last column shows the enlarged details indicated by the digits. Digits "1", "2", "3", and "4" represent predictions of ground truth, FFL, PolyWorld, and PolyBuilding, respectively. The white circles indicate missed and false detected buildings. The white arrow in the third image of PolyWolrd predictions points out a miss detected vertex.
Best viewed in color.}
\label{fig-sota}
\end{figure*}

We also visualize some prediction results for qualitative comparison among PolyBuilding, FFL, and PolyWorld as shown in Figure \ref{fig-sota}. PolyBuilding successfully detects all buildings, regardless of their different sizes.
However, PolyWorld miss detects some buildings, and FFL falsely detects some other objects as buildings. We mark the missed and false detections with white circles in Figure \ref{fig-sota}.
As shown in the third column of Figure \ref{fig-sota}, several missed detected buildings are in PolyWorld predictions.
For example, a medium-sized and a large-sized building in the last image are miss detected by PolyWorld.
As for FFL, it detects many other small objects as buildings, as shown in the second column of Figure \ref{fig-sota}. For example, it detects a small square ground area as a building in the first image and even detects a car as a building in the third image.
By contrast, PolyBuilding detects all building instances correctly in these example images. The superior instance-level detection performance explains our model's high AP and AR values, as shown in Table \ref{tab-sota}.
In addition, FFL and PolyWorld struggle to predict accurate vertex locations when building corners are occluded. 
For example, as shown in the first row of Figure \ref{fig-sota}, a part of the building in the upper left corner of the image is occluded by vegetation. We mark the occluded buildings using digits and enlarge them in the last column of Figure \ref{fig-sota}.
As can be seen, except PolyBuilding, all the other compared methods (FFL and PolyWorld) can not predict correct locations for the occluded corners.
In addition, there is a similar case in the last image. As indicated by the digits, two out of four corners are occluded by vegetation. In this example, FFL can not locate the occluded corners similarly. PolyWorld miss detects the whole building. Only our model predicts it correctly.
The reason can be analyzed from how the models predict building corners.
FFL is a segmentation-based method, and the building segmentation mask is not complete in the case of occlusion. Therefore, during polygonization, the vertices transferred from segmentation contours are in wrong locations.
PolyWorld obtains the locations of building corners by vertex detection (segmentation), thus occluded corners may be miss detected. This can also explain the miss detected large building in the last image. The whole building is missed may be because all the four corners are not detected due to the low color contrast with the background.
There is also a miss detected corner in the third image of the PolyWorld predictions, indicated by the white arrow.
By contrast, PolyBuilding can successfully handle the occlusion case since it regresses the vertex locations directly, which supports it in inferring the locations of occluded vertices according to the locations of other vertices.
What's more, PolyBuilding can better handle round buildings. For example, as indicated by the digits in the second row of Figure \ref{fig-sota}, FFL predicts four corners for the round building, and PolyWorld miss detects it due to its indistinct vertex feature. However, PolyBuilding predicts a circle of vertices along the building boundary, approximating the ground truth best.
The qualitative results further demonstrate the superiority of our PolyBuilding model.

\section{Discussion}
\subsection{End-to-End Polygonal Building Extraction Methods}
As mentioned before in the related work section, previous end-to-end polygonal building extraction methods can be mainly classified into two categories. One category of methods \cite{polymapper,polymapper-follower,curvegcn-based} are based on object detection. They first detect building bounding boxes using an object detector and then predict polygons within the bounding boxes. The polygon prediction can be achieved using an RNN to predict the corners sequentially or a GCN to predict the vertices simultaneously. The CNN-RNN or CNN-GCN paradigm are multi-stage and difficult to train and inference.
The other category of methods circumvents object detection for buildings \cite{polyworld}. They first predict all building corners in the input image through vertex segmentation and then connect the extracted vertices to form the polygons according to a learned connection matrix. This category of methods is more concise, but there are also some limitations, except the aforementioned miss-detection issue. For example, the connection between extracted vertices are determined by the learned matrix, which can not guarantee the generated polygons to be closed and non-intersected, which are another two important properties for assessing polygonal buildings.
Our proposed PolyBuilding method does not fall into either of the two categories since it regresses the building polygon coordinates directly. The bounding boxes are also predicted in our model, but they are only used for the matching between ground truth and predicted instances during the bipartite matching procedure. Actually, the bounding box prediction can be eliminated if we use the polygon to perform the matching, which would results in a simpler model. We may try this in our future research.
In addition, the generated polygons by PolyBuilding are guaranteed to be closed since the start and end points of the predicted vertex sequence will be connected. As for intersection, although there is no guarantee for non-intersection of our model, we do not observe any intersections in the generated polygons, which may be due to the dense vertex supervision signals provided by the uniform sampled ground truth.

\subsection{Analysis of Failure Cases}

\begin{figure}[!t]
\centering
\includegraphics[width=8.5cm]{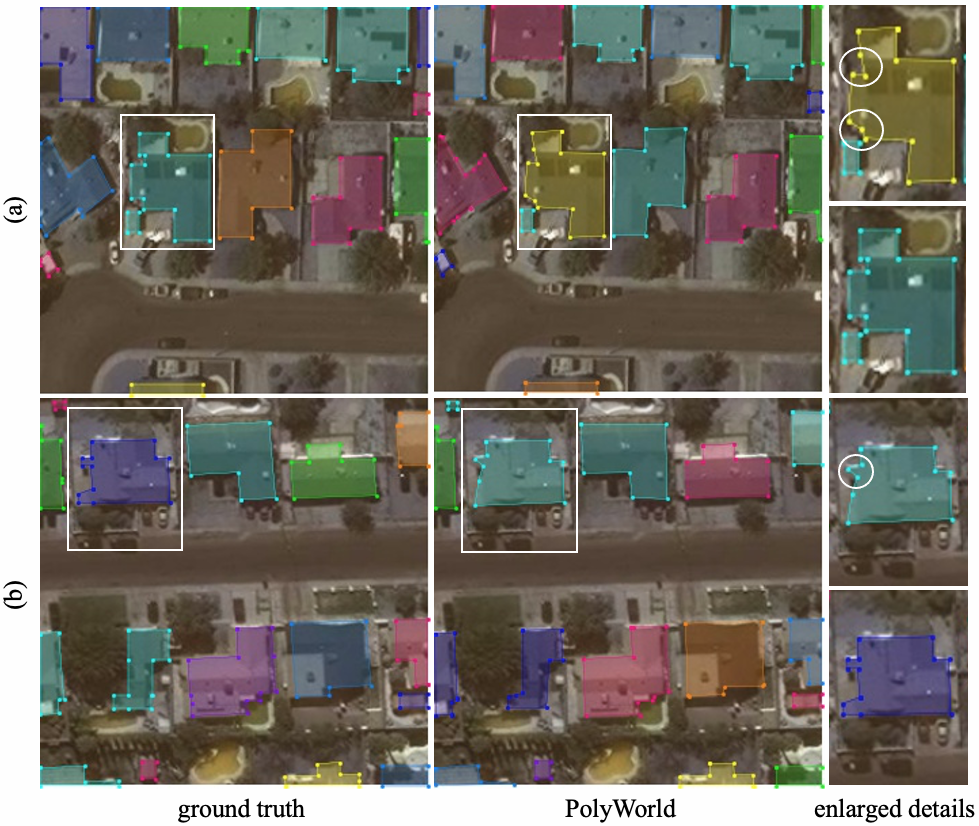}
\caption{Visualization of some cases that our PolyBuilding fails. (a) and (b) show two cases in which we enlarge the polygons indicated by the rectangles in the last column. The circles indicate some acute and obtuse angles generated by PolyBuilding.}
\label{fig-failure-cases}
\end{figure}

Although our proposed PolyBuilding achieves new state-of-the-art performance on the CrowdAI dataset, there are still some failure cases. As illustrated in Fig \ref{fig-failure-cases}, two prediction results of PolyBuilding is visualized. The generated polygons in these two samples have acute and obtuse angles, which are indicated by the circles. The reasons for the two failure cases are different. Case (a) is caused by inaccurate prediction of vertex locations. Our model does not adopt any geometric constraint loss to encourage the corners to be orthogonal. This may be alleviated by adding the angle difference loss as in \cite{polyworld}. The sharp angle in case (b) is caused by the NMS refinement operation. As can be seen, the sharp angle occurs on a short edge. If the two corners of the edge are not local maximum in the corner score sequence, NMS would filter them out and only keep one vertex with the highest score for the edge, thus causing the sharp angle.
However, in most cases, the corners achieve the maximum confidence score in their neighborhood, and NMS can effectively filter out the non-maximum vertices and reduce redundancy near the corners.

\section{Conclusion}
In this paper, we propose a fully end-to-end model (PolyBuilding) for polygonal building extraction, which predicts vector representation of buildings from the remote sensing imagery directly.
Motivated by the similarity of box detection and polygon prediction, we extend the Deformable DETR \cite{deformable-detr} to further regress polygon coordinates besides the bounding boxes by adding the polygon regression head and the corner classification head.
The polygon regression head predicts a set of vertex coordinate sequences for building polygons, and the corner classification head predicts the corresponding score sequences measuring the probability of the vertices being corners.
In order to unify the input and output shape of multiple polygon instances, we design a uniform sampling encoding scheme to generate the ground truth with the same vertex number for all buildings.
During inference, the corner scores and NMS are leveraged to reduce the vertex redundancy of the generated polygons.
Moreover, we introduce a two-phase strategy to facilitate the training of the model by decomposing the tasks of location learning and corner classification.
Extensive experiments on the CrowdAI dataset demonstrate the superiority of our proposed PolyBuilding in terms of pixel-level coverage, instance-level detection accuracy, and geometry-level properties.

\bibliographystyle{IEEEtran}
\bibliography{bare_jrnl_new_sample4}

\vfill

\end{document}